\def\year{2022}\relax
\documentclass[letterpaper]{article} % DO NOT CHANGE THIS
\usepackage{aaai}  % DO NOT CHANGE THIS
\usepackage{times}  % DO NOT CHANGE THIS
\usepackage{helvet}  % DO NOT CHANGE THIS

\usepackage{courier}  % DO NOT CHANGE THIS
\usepackage[hyphens]{url}  % DO NOT CHANGE THIS
\usepackage{graphicx} % DO NOT CHANGE THIS
\usepackage{natbib}  % DO NOT CHANGE THIS AND DO NOT ADD ANY OPTIONS TO IT
\usepackage{caption} % DO NOT CHANGE THIS AND DO NOT ADD ANY OPTIONS TO IT
\DeclareCaptionStyle{ruled}{labelfont=normalfont,labelsep=colon,strut=off} % DO NOT CHANGE THIS
\usepackage{algorithm}
\usepackage{algpseudocode}
\usepackage{amsmath}
\usepackage{mathrsfs}
\usepackage{makecell}
\usepackage{amsbsy}
\usepackage{bibunits}
\usepackage{amssymb}
\usepackage{changes}

\usepackage{newfloat}
\usepackage{listings}
\floatstyle{ruled}
\newfloat{listing}{tb}{lst}{}
\floatname{listing}{Listing}
\usepackage{subfig}
\usepackage{multirow}
\usepackage{booktabs}
\usepackage{amsthm}

\theoremstyle{definition}
\newtheorem*{remark}{Remark}
\newtheorem{definition}{Definition}[section]
\newcommand{\Tau}{\mathcal{T}}
\begin{document}
\title{{InfoLM}: A New Metric to Evaluate Summarization \\ \& Data2Text Generation}
\author{
Pierre Colombo$^{1}$\thanks{Work done while Pierre was funded by IBM \& Telecom Paris},  Chlo\'e Clavel$^2$,Pablo Piantanida$^1$
  }
  \affiliations{
    $^1$Laboratoire des Signaux et Systèmes (L2S), CentraleSupelec CNRS Universite Paris-Saclay \\
  $^2$T\'el\'ecom ParisTech, Universit\'e Paris Saclay \\
  {pierre.colombo@centralesupelec.fr} \\
  {chloe.clavel@telecom-paris.fr} \\
  {pablo.piantanida@centralesupelec.fr}}
\maketitle

\begin{abstract}
Assessing the quality of natural language generation systems through human annotation is very expensive. Additionally, human annotation campaigns are time-consuming and include non-reusable human labour. In practice, researchers rely on automatic metrics as a proxy of quality. In the last decade, many string-based metrics (e.g., BLEU) have been introduced. However, such metrics usually rely on exact matches and thus, do not robustly handle synonyms. In this paper, we introduce InfoLM a family of untrained metrics that can be viewed as a string-based metric that addresses the aforementioned flaws thanks to a pre-trained masked language model. This family of metrics also makes use of information measures allowing the adaptation of InfoLM to various evaluation criteria. Using direct assessment, we demonstrate that InfoLM achieves statistically significant improvement and over 10 points of correlation gains in many configurations on both summarization and data2text generation.
\end{abstract}

\section{Introduction}
A plethora of applications of natural language processing (NLP) performs text-to-text transformation \cite{mellish1998evaluation,belz2006comparing,specia2018quality}. Given an input, these systems are required to produce an output text that is coherent, readable and informative. Due to both high annotation costs and time, researchers tend to rely on automatic evaluation to compare the outputs of such systems. 
Reference-based automatic evaluation relies on comparing a candidate text produced by the NLG system and one or multiple reference texts (‘gold standard’) created by a human annotator. Generic automatic evaluation of NLG is a huge challenge as it requires building a metric that evaluates the similarity between a candidate and one or several gold-standard reference texts. However, the definition of success criteria is task-specific: as an example, evaluation of text summarization focuses on content, coherence, grammatically, conciseness,  and readability \cite{mani2001automatic,colombo2022best}, whereas machine translation focuses on fidelity, fluency and adequacy of the translation \cite{hovy1999toward} and data2text generation \cite{gardent2017creating} consider criteria such as data coverage, correctness and text structure. 

Automatic text evaluation metrics fall into two categories: metrics that are trained to maximise their correlations using human annotation (\textit{e.g.}, \texttt{BLEND} \cite{ma2017blend}) and untrained metrics (\textit{e.g.}, \texttt{BLEU} \cite{bleu}, \texttt{ROUGE} \cite{lin-2004-rouge}, \texttt{BERTSCORE} \cite{zhang2019bertscore}, \texttt{DepthScore} \cite{staerman2021depth}, \texttt{BaryScore} \cite{colombo2021automatic},   \texttt{MOVERSCORE} \cite{zhao2019moverscore}). In this work, we focus on untrained metrics as trained metrics may not generalize well to new data (existing labelled corpora are of small size). Two categories of untrained metrics can be distinguished: word or character based-metrics that compute a score based on string representation and  embedding-based metrics that rely on a continuous representation. String-based metrics (\textit{e.g.}, \texttt{BLEU}) often fail to robustly match paraphrases \cite{reiter2009investigation} as they mainly focus on the surface form as opposed to embedding-based metrics relying on continuous representations.

In this paper, we introduce \texttt{InfoLM} a family of new untrained metrics to evaluate text summarization and data2text generation. At the highest level \texttt{InfoLM} key components include: (1) a pre-trained masked language model (\texttt{PMLM}) that is used to compute two discrete probability distributions over the vocabulary. They represent the probability of observing each token of the vocabulary given the candidate and the reference sentence, respectively. (2) A contrast function $\mathcal{I}$ that is used to measure the dissimilarity between aforementioned probability distributions. \texttt{InfoLM} differs from existing BERT-based metrics (\textit{e.g.} \texttt{BERTSCORE}, \texttt{MOVERSCORE}) as it directly relies on the \texttt{PMLM} which outputs discrete probability distributions. Thus \texttt{InfoLM} does neither require to arbitrarily select one or several specific layers ( \textit{e.g.} \texttt{BERTSCORE} relies on the 9th layer for \texttt{bert-base-uncased}), nor involves selecting arbitrary aggregations technics  (\textit{e.g.} see \texttt{MOVERSCORE}). As \texttt{InfoLM} relies on statistics on tokens it can also be seen as a string-based metric. However, it does not suffer from common pitfalls of string-based metrics (\textit{e.g.} synonyms, need of an exact string-match) as the \texttt{PMLM} also allows ones to assign a high score to paraphrases and to capture distant dependencies.
\\\noindent \textbf{Contributions}
Our contributions are summarized below:
\\(1) \textit{A set of novel metrics to automatically evaluate summarization and data2text generation}. In this work, we introduce \texttt{InfoLM} which overcomes the common pitfall of string matching metrics and does not require to select a layer, nor to rely on a arbitrary aggregation function. \texttt{InfoLM} combines a pre-trained model and a contrast function denoted by $\mathcal{I}$ between two discrete probability distributions. We explore the use of different choices of contrast functions such as $f$-divergences, $\mathcal{L}_p$ distances or Fisher-Rao distances.
    %Numerical Results
\\(2) \textit{Tasks}. First, we demonstrate on both summarization and data2text that \texttt{InfoLM} is better suited than concurrent metrics. A comparison is conducted, using multiple correlation measures with human judgment both at the text and system level. Second, we dissect \texttt{InfoLM} to better understand the relative importance of each component (\textit{e.g.} calibration, sensibility to the change of information measures).

\section{Problem Statement and Related Work}
In this section, we start by introducing notations and formulate the problem of both evaluating text generation and  metrics. Then, we identify and present the most relevant related work and the existing approaches for the studied tasks.

\subsection{Problem Statement}

\textbf{NLG evaluation.} Given a dataset $\mathcal{D} = \{\pmb{x_i}, \{\pmb{y_i^s},h(\pmb{x_i},\pmb{y_i^s})\}_{s=1}^S \}_{i=1}^N$ where $\pmb{x}_i$ is the $i$-th reference text; $\pmb{y}_i^s$ is the $i$-th candidate text generated by the $s$-th NLG system; $N$ is the number of texts in the dataset and $S$ the number of systems available. The vector $\pmb{x_i} = ({x_1,\cdots,x_M})$ is composed of M tokens (\textit{e.g.}, words or subwords) and $\pmb{y_i^s} = ({y_1^s,\cdots,y_L^s})$ is composed of L tokens\footnote{The reference and candidate text can be composed of several sentences as it is the case in summarization.}. The set of tokens (vocabulary) is denoted as $\Omega$, $\mathbf{T}$ denotes the set of possible texts.  $h(\pmb{x_i},\pmb{y_i^s}) \in \mathcal{R}^+$ is the score associated by a human annotator to the candidate text $\pmb{y}_i^s$ when comparing it with the reference text $\pmb{x_i}$. We aim at building an evaluation metric $f$ such that $f(\pmb{x_i} ,\pmb{y_i})\in \mathbb{R}^{+}$. 

\noindent\textbf{Evaluating evaluation metrics.} To assess the relevance of an evaluation metric $f$, correlation with human judgment is considered to be one of the most important criteria \cite{koehn2009statistical,chatzikoumi2020evaluate}. Debate on the relative merits of different correlations for the evaluation of automatic metrics is ongoing but classical correlation measures are Pearson \cite{leusch2003novel}, Spearman \cite{melamed2003precision} or Kendall \cite{kendall1938new} tests. Two meta-evaluation strategies are commonly used: (1) text-level correlation or (2) system-level correlation. Formally, the text-level correlation $C_{t,f}$ is computed as follows: 
\begin{equation} \label{se}
\begin{split}
    C_{t,f} &\triangleq  \frac{1}{N} \sum_{i=1}^N K ( \mathbf{F}^t_i, \mathbf{H}^t_i ),  
\end{split}
\end{equation} 
where $\mathbf{F}_i = \big{[}f(\pmb{x_i},\pmb{y_i^1}),\cdots,f(\pmb{x_i},\pmb{y_i^S})\big{]}$ and $\mathbf{H}_i =\big{[}h(\pmb{x_i},\pmb{y_i^1}),\cdots,h(\pmb{x_i},\pmb{y_i^S})\big{]}$ are the vectors composed of scores assigned by the automatic metric $f$ and the human respectively.
and $K:\mathbb{R}^N \times\mathbb{R}^N \rightarrow [-1,1] $ is the chosen correlation measure (\textit{e.g.}, Pearson, Kendall or Spearman).
Similarly, the system level correlation $C_{sy,f}$ is obtained by

\begin{align} \label{sy}
 C_{sy,f} &\triangleq  K ( \mathbf{F}^{sy}, \mathbf{H}^{sy} ),\\
\mathbf{F}^{sy} & = \left[\frac{1}{N} \sum\limits_{i=1}^N f(\pmb{x_i},\pmb{y_i^1}),\dots, \frac{1}{N}\sum\limits_{i=1}^N f(\pmb{x_i},\pmb{y_i^S})\right]\nonumber \\
\mathbf{H}^{sy} & = \left[\frac{1}{N}\sum\limits_{i=1}^N h(\pmb{x_i},\pmb{y_i^1}),\dots,\frac{1}{N}\sum\limits_{i=1}^N h(\pmb{x_i},\pmb{y_i^S})\right],\nonumber 
\end{align}
where the latter are the vectors composed of the averaged scores assigned by $f$ and the human, respectively.
For the significance analysis, we follow \citet{graham2014testing} and use a William test to validate a significant improvement for dependent correlations \cite{steiger1980tests}. 

\subsection{Existing Metrics}
\subsubsection{String-based Metrics}
Two types of string-based metrics exist: \emph{N-Grams matching} and \emph{Edit distance-based} metrics. \emph{N-Grams matching metrics} count the number of N-grams in common between the candidate text and the reference text. The three most-used metrics are \texttt{BLEU}, \texttt{ROUGE} and \texttt{METEOR} \cite{banerjee2005meteor}. If no N-gram is in common between the input text candidate and the reference, these metrics fail to produce meaningful scores. The second category of metrics gathers \emph{edit distance-based metrics}. They measure the number of basic operations such as edit/delete/insert to measure semantic equivalence. Variants include \texttt{TER} \cite{snover2006ter}, \texttt{CDER} \cite{leusch-etal-2006-cder}, \texttt{EED} \cite{stanchev2019eed}, \texttt{CHARACTER} \cite{wang2016character}. Edit distance-based metrics do not handle synonyms and focus on surface form. \texttt{InfoLM} can be seen as string-based but do not suffer from the aforementioned matching problem and can handle synonyms as it relies on a \texttt{PMLM}.
\subsubsection{Embedding-based Metrics} Another class of metrics relies on word embeddings. These metrics either use static words embeddings such as word2vec \cite{word2vec} or contextualized embeddings such as ELMO \cite{elmo}, BERT \cite{bert} and its variants \cite{sanh2019distilbert,liu2019roberta}. Among the most popular metrics, we can mention \texttt{MOVERSCORE}, \texttt{BERTSCORE}, \texttt{WMD} \cite{kusner2015wmd} , \texttt{WMDO} \cite{chow-etal-2019-wmdo}. Different from these approaches \texttt{InfoLM} relies on a language model and work with discrete probability distributions instead of continuous representations. 
\subsubsection{Learning-based Metrics} 
Various trained metrics have been proposed such as \texttt{BEER}, \texttt{BEND}, \texttt{RUSE}, \texttt{CIDER} \cite{vedantam2015cider}. These methods rely on train/dev/test sets composed of human evaluations. \texttt{InfoLM} does not require any training step and relies on a frozen \texttt{PMLM}.
\subsubsection{\texttt{PMLM} as a Metric.} To the best of our knowledge, using a \texttt{PMLM} (\textit{i.e}, without further training) as a reference-based automatic metric remains overlooked. The closest use we found was to rely on autoregressive models, such as GPT-2 \cite{radford2019language}, to compute the generated sentence perplexity and assess its fluency. Researchers mainly focused on the use of the learnt embedding of the \texttt{PMLM}. However, it remains an open question to find a reliable layer aggregation mechanism (\textit{e.g}, \texttt{BERTSCORE} arbitrary selects a layer based on a chosen dataset, \texttt{MOVERSCORE} uses the 5 last layers). \texttt{InfoLM} addresses this aggregation issue by relying on the \texttt{PMLM}.

% In text generation (\textit{e.g.}, style transfer, news generation), LM is (optionally) fine-tuned to measure perplexity and assess the fluency of the generated sentences.

% where $\pmb{M}\in\{1,\dots, M\}$ denotes a random vector indicating the selected set of masked positions and $p_{\mathbf{\Omega} | \mathbf{X}}(\cdot  | [\pmb{x}]) $ represents the output of the softmax layer of the considered LM.

\subsection{Masked Language Modeling}
Language models based on masked language pre-training objectives \citet{bert,liu2019roberta} aim at reconstructing a corrupt version $[\pmb{x}]$ of an input text $\pmb{x}$ by minimizing a cross-entropy loss. This corrupted context corresponds to a "local view" view of the sentence. To ensure fair comparison, we do not use existing alternatives (\textit{e.g.} GPT-2 based models \cite{radford2019language}) as concurrent works \cite{zhang2019bertscore,zhao2019moverscore} rely on \texttt{PMLM}.

\section{InfoLM}
In this section, we first introduce a novel family of metrics called \texttt{InfoLM} and then detail the different components of these novel metrics. 
\\\noindent\textbf{Notations} We denote by $\theta \in \Theta$, the  parameter of the \texttt{PMLM},  $\Tau$ its temperature and $\mathcal{I} : [0,1]^{|\mathbf{\Omega}|} \times [0,1]^{|\mathbf{\Omega}|}$ an information measure (see \ref{ssec:information_measures}) which quantifies the similarity between two discrete distributions.

\subsection{Motivations \& Definitions} 
A \texttt{PMLM} has learnt the empirical distribution of a large text corpus. Given a text $\pmb{x}$, the corrupted context with a mask at position j is denoted $[\pmb{x}]^j$, the LM predicts a distribution $p_{\mathbf{\Omega} | \mathbf{T}}(\cdot |[\pmb{x}]^j; \theta; \Tau)$ over the vocabulary $\mathbf{\Omega}$ given the masked context. As an example, for a masked input sentence $[\pmb{x}]^1 = \text{“The [MASK] was delicious”}$, a pretrained model could place high probabilities on tokens “food”, “meal” and low probability on "the". It is worth noting that $p_{\mathbf{\Omega} | \mathbf{T}}(\cdot |[\pmb{x}]^1; \theta; \Tau)$ represents the probability of observing each token of the vocabulary given the masked input $[\pmb{x}]^1$.

\begin{definition}[\textit{Equivalence for masked contexts}]\label{def:masked_contexts} Given $\mathcal{I}$, two masked contexts $[\pmb{x}]^j$, $[\pmb{y}]^k$ from input texts $\pmb{x}$, $\pmb{y}$, with masks at positions j and k respectively, are equivalent (denoted $[\pmb{x}]^j \overset{\mathcal{I}}{\sim} [\pmb{y}]^k$) if the two predicted discrete distributions given by the \texttt{PMLM}, namely $p_{\mathbf{\Omega} | \mathbf{T}}(\cdot |[\pmb{x}]^j; \theta; \Tau)$ and $p_{\mathbf{\Omega} | \mathbf{T}}(\cdot |[\pmb{y}]^k; \theta; \Tau)$, are similar. Formally, $[\pmb{x}]^j \overset{\mathcal{I}}{\sim} [\pmb{y}]^k$ if $\mathcal{I}\left[p_{\mathbf{\Omega} | \mathbf{T}}(\cdot |[\pmb{x}]^j; \theta; \Tau),p_{\mathbf{\Omega} | \mathbf{T}}(\cdot |[\pmb{y}]^k; \theta; \Tau)\right] \approx 0$.
\end{definition}
\begin{remark} We have the intuition that two similar sentences will share several pairs of equivalent masked contexts. At this point, we make no claim on the relationship between equivalence and the masked context similarity. 
\end{remark}

\noindent In this work, we make the hypothesis that two similar sentences $\pmb{x},\pmb{y}$ will share multiple equivalent masked contexts. However, pairwise comparisons of all the pairs of individual masked contexts are prohibitively expensive ($\mathcal{O}(L \times M)$ comparisons) when considering long texts. Motivated by efficiency, we instead propose to work with two "global views" of the sentences that are well-formed probability distributions and are obtained through the aggregation of individual \texttt{PMLM} predictions. Aggregated probabilities for $\pmb{x}$ and $\pmb{y}$ are denoted $p_{\mathbf{\Omega} | \mathbf{T}}(\cdot |\pmb{x}; \theta; \Tau)$ and $p_{\mathbf{\Omega} | \mathbf{T}}(\cdot|\pmb{y}; \theta; \Tau)$ respectively. 
\begin{definition}[\textit{Similarity for texts}]\label{def:similar_texts} Given $\mathcal{I}$, two texts $\pmb{x},\pmb{y}$ are said to be similar (denoted $\pmb{x} \overset{\mathcal{I}}{\approx} \pmb{y}$) if  $\mathcal{I}\left[p_{\mathbf{\Omega} | \mathbf{T}}(\cdot |\pmb{x}; \theta; \Tau),p_{\mathbf{\Omega} | \mathbf{T}}(\cdot |\pmb{y}; \theta; \Tau)\right] \approx 0$ where $p_{\mathbf{\Omega} | \mathbf{T}}(\cdot |\pmb{x}; \theta; \Tau)$ and $p_{\mathbf{\Omega} | \mathbf{T}}(\cdot|\pmb{y}; \theta; \Tau)$ denotes the aggregated individual masked context predictions.
\end{definition}

\subsection{InfoLM}
\subsubsection{Overview}
\texttt{InfoLM} uses the notion of similarity given in \ref{def:similar_texts}. Given a reference text $\pmb{x_i}$ together with a candidate text $\pmb{y_i^s}$, \texttt{InfoLM} recursively masks each token position of both $\pmb{x_i}$ and $\pmb{y_i^s}$ to obtain individual masked contexts. By relying on a \texttt{PMLM}, \texttt{InfoLM} predicts one distribution for each individual masked contexts. The resulting distributions are then averaged (we refer to this operation "bag of distributions") to obtain $p_{\mathbf{\Omega} | \mathbf{T}}(\cdot |\pmb{y_i^s}; \theta; \Tau)$ and $p_{\mathbf{\Omega} | \mathbf{T}}(\cdot|\pmb{x_i}; \theta; \Tau)$. The final step involves comparing two well formed discrete probability distributions $p_{\mathbf{\Omega} | \mathbf{T}}(\cdot |\pmb{y_i^s}; \theta; \Tau)$ and $p_{\mathbf{\Omega} | \mathbf{T}}(\cdot|\pmb{x_i}; \theta; \Tau)$ through a measure of information $\mathcal{I}$. \texttt{InfoLM} writes as:
{\small
\begin{equation}\label{eq:infolm}
 \scalebox{1}[1]{ \texttt{InfoLM}}\hspace{-.2em}\left(\pmb{x_i},\pmb{y_i^s}\right) \hspace{-.2em}\triangleq\hspace{-.2em}\mathcal{I}\big[p_{\mathbf{\Omega} | \mathbf{T}}(\cdot |\pmb{x_i}; \theta; \Tau),p_{\mathbf{\Omega} | \mathbf{T}}(\cdot|\pmb{y_i^s}; \theta; \Tau)\big].\hspace{-.2em}
\end{equation}}
\begin{remark} It is worth to emphasize that 
$p_{\mathbf{\Omega} | \mathbf{T}}(\cdot|\pmb{x_i}; \theta; \Tau)$ and $p_{\mathbf{\Omega} | \mathbf{T}}(\cdot|\pmb{y_i^s}; \theta; \Tau)$ are two well formed discrete probability distributions. They represent the probability of observing each token of the vocabulary given the candidate and the reference sentence, respectively. 
\end{remark}

\subsubsection{Aggregation Procedure} Rare tokens can be more indicative of text similarity than common tokens \cite{banerjee2005meteor}. Thus, for the aggregation of the individual masked contexts, we propose to compute a weighted "bag of distributions" where the weights are normalized measures of the importance of each token. In practice,  $p_{\mathbf{\Omega} | \mathbf{T}}(\cdot|\pmb{x}; \theta; \Tau)$ and  $p_{\mathbf{\Omega} | \mathbf{T}}(\cdot| \pmb{y_i^s}; \theta; \Tau)$ write as:
\begin{equation*}
   \begin{split}
  p_{\mathbf{\Omega} | \mathbf{T}}(\cdot|\pmb{x}; \theta; \Tau)& \triangleq  \sum_{k=1}^M \gamma_k \times p_{\theta}(\cdot|[\pmb{x}]^k; \theta; \Tau), \\
  p_{\mathbf{\Omega} | \mathbf{T}}(\cdot |\pmb{y_i^s}; \theta; \Tau) &   \triangleq  \sum_{k=1}^N \Tilde{\gamma}_k \times p_{\theta}(\cdot|[\pmb{y_i^s}]^k; \theta; \Tau),
    \end{split}\end{equation*}
where $\Tilde{\gamma}_k$ and $\gamma_k$ are measures of the importance of the k-th token in the candidate and reference text, respectively, i.e., satisfying  $\sum_{j=1}^M \gamma_j=\sum_{j=1}^N \tilde{\gamma}_j=1$. These are  computed using the inverse document frequency scores determined at the corpus level \cite{zhao2019moverscore,kusner2015wmd}.

\subsubsection{LM Calibration.} Modern deep neural networks are overconfident \cite{guo2017calibration}. To re-calibrate language models several techniques have been proposed (\textit{e.g} temperature scaling \cite{platt1999probabilistic}). Here, we choose to study how calibration affects \texttt{InfoLM} by relying on temperature\footnote{When $\Tau \rightarrow 0$ one token receive all the probability mass, and when $\Tau \rightarrow \infty$, the probability becomes uniform.} scaling motivated by simplicity and speed. 
\subsection{Information Measures}\label{ssec:information_measures}
In this work, we focus on comparing a pair of discrete probability distributions through information measures (see \citet{basseville2013divergence} for an exhaustive study). We rely on two types of information measures: \emph{divergences} and \emph{distances}. The divergence is a measure of dissimilarity that is always positive or equal to zero if (and only if) the two considered distributions are strictly identical. We call distance, a function that is symmetric, positive, respects the triangle inequality and is equal to zero if (and only if) the two considered distributions are strictly identical. We will use information measures that belong to either Csiszar $f$-divergences or that are distances.

\begin{table*}[]
    \centering
   \resizebox{.9\textwidth}{!}{ \begin{tabular}{cc|c|c}\hline\hline
     Name & Notation  & Domain & Expression   \\\hline\hline
%          \makecell{$\alpha$-Rényi divergence \\ \cite{renyi1961measures}} &     $\mathcal{D}_R^\alpha $   &    $\alpha \in $ &   $\frac{1}{\alpha - 1 }\log \sum p_i^{1-\alpha} q_i^\alpha$ \\
     \makecell{$\alpha$-divergence \\ \cite{csiszar1967information}} &  $\mathcal{D}_\alpha $  & $\alpha \not\in \{0,1\}$ & $\frac{1}{\alpha(\alpha - 1)}(1 -  \sum q_i^{1-\alpha} p_i^\alpha)$  \\
     \makecell{$\gamma$ divergence\\ \cite{fujisawa2008robust}} &   $\mathcal{D}_\gamma^\beta$  & $\beta \not\in \{0,-1\}$ &   $ \frac{1}{\beta (\beta + 1)} \log \sum p_i^{\beta + 1} + \frac{1}{\beta + 1} \log \sum q_i^{\beta + 1} -  \frac{1}{\beta}\log \sum p_i q_i^\beta$ \\
          \makecell{AB Divergence\\ \cite{cichocki2011generalized}}  &     $\mathcal{D}_{sAB}^{\alpha,\beta}$ & \makecell{ $(\alpha,\beta) \in (\mathbb{R}^*)^2$ \\  $\beta + \alpha \neq 0$ }& $ \frac{1}{\beta (\beta + \alpha)} \log \sum p_i^{\beta + \alpha} + \frac{1}{\beta + \alpha} \log \sum q_i^{\beta + \alpha} -  \frac{1}{\beta} \log\sum p_i^\alpha q_i^\beta$ \\\hline
    $\mathcal{L}_1$ distance & $\mathcal{L}_1$ & & $\sum |p_i - q_i|$ \\
    $\mathcal{L}_2$ distance & $\mathcal{L}_2$ & & $\sqrt{\sum (p_i - q_i)^2}$ \\
    $\mathcal{L}_\infty$ distance & $\mathcal{L}_\infty$ & & $\max_i|p_i - q_i|$\\
    Fisher-Rao distance & $R$ & & $\frac{2}{\pi}\arccos{\sum \sqrt{p_i \times q_i}}$\\\hline\hline
    \end{tabular}}
    \caption{Expression of the divergences (upper group) and distance between two positives measures $\mathbf{p} = (p_1,\cdots,p_N)$ and $\mathbf{q} = (q_1,\cdots,q_N)$ as well as the definition domain (see \cite{regli2018alpha}). We omit the index in the summations.}
    \label{tab:contrast_measures}
\end{table*}

\subsubsection{Divergence Measures}
Various divergence measures have been proposed for a large variety of applications \cite{basseville2013divergence}. The full expression of the studied divergences can be found in \ref{tab:contrast_measures}. We focus here on three families of divergences $\alpha$ Divergences, $\gamma$ Divergences and $AB$ Divergences.  Note that there exist other families of divergences such as Bregman divergence \cite{bregman1967relaxation}, $\beta$ divergences \cite{basu1998robust}, Chernoff divergence \cite{chernoff1952measure}.
\\\textbf{$\alpha$-Divergences. } 
This divergence was introduced by \citet{renyi1961measures} and are a special case of the $f$-divergences \cite{csiszar1967information}.  They are widely used in variational inference \cite{li2016r} and closely related to Rényi divergences but are not a special case.  
From \ref{tab:contrast_measures} we note special cases of $\alpha$-Divergences: (i) Kullback-Leiber (KL) is recovered by letting $\alpha \rightarrow 1$, (ii) Hellinger distance \cite{hellinger1909neue} follows by choosing $\alpha = 0.5$.
For this family, $\alpha$ weights the influence of $\frac{p}{q}$.
\\\textbf{$\gamma$-Divergences.} This divergence has been introduced by \cite{fujisawa2008robust} as a scale-invariant modification of the robust $\beta$-divergences.\footnote{In our setting we work we normalised distributions, thus scale invariance is not a mandatory property. It is worth mentioning as it could cause practical issues when optimising our metric.} For the $\gamma$ divergences the parameter $\beta$ is used to control the importance of the element of small probabilities (\textit{e.g.}, outliers in some scenarios, tokens with low probability in our case). If $\beta > 1$, the importance of large $q_i$ is reduced which gives more weights to the outliers. Special cases include the ${L}_2$ distance (i.e., $\beta=2$) and KL divergence (i.e., $\beta \rightarrow 1$). 
\\\textbf{$AB$-Divergences.} The family of $AB$-divergences is flexible and allows to respectively control the mass coverage or the robustness. \citet{cichocki2011generalized} propose to use $AB$ divergences. As can be seen in \ref{tab:contrast_measures} these divergences have two parameters $\alpha,\beta$. It allows to tune the mass coverage and the robustness independently. The $\beta$-divergence is obtained by choosing $\alpha = 1, \beta \in \mathbb{R}$.
\\\noindent\textbf{From information divergences to discrimination.} For our application, we would like to produce a metric between two texts regardless of the source (system or human). Thus we are interested in symmetric divergence: such divergences are called discrimination. To obtain discrimination two tricks are commonly applied either the Jeffrey's symmetrization, which averages $\textrm{KL}(p\|q)$ and  $\textrm{KL}(q\|p)$), or the Jensen's symmetrization, which averages $\textrm{KL}(p\|\frac{p+q}{2})$ and  $\textrm{KL}(q\|\frac{p+q}{2})$. We choose to use Jeffreys symmetrization as it does not require computing $\frac{p+q}{2}$. The symmetric $KL$ with Jeffrey's symmetrization is denoted $JS$.
\subsubsection{Distances}
%In addition to divergences, we consider t study two distances $L_p$ and Rao. 
\textbf{$\mathcal{L}_p$ distances.} The $\mathcal{L}_p$ distances $p \in \mathbb{R}_{+}$ can be used to measure the similarity between two distributions and we restrict ourselves to $p \in \{1,2,+\infty\}$.
\\\textbf{Fisher-Rao distance.} The Fisher-Rao distance ($R$) represents the Geodesic Distance \cite{rao1987differential} between two distributions. Interestingly, this distance remains overlooked in the ML community but has been recently used to achieve robustness against adversarial attacks \cite{marine_rao}.

\subsubsection{Connection with String Matching Metrics.}\label{ssec:string_matching}
We adopt the following notations $p_{\mathbf{\Omega} | \mathbf{T}}(\cdot |\pmb{x_i}; \theta; \Tau) = [p_0,\cdots,p_{|\Omega|}]$ and $p_{\mathbf{\Omega} | \mathbf{T}}(\cdot|\pmb{y_i^s}; \theta; \Tau) = [q_0,\cdots,q_{|\Omega|}]$. First, let us consider two texts $\pmb{x_i},\pmb{y_i^s}$ such that  $\pmb{x_i} \overset{\mathcal{I}}{\approx} \pmb{y_i^s}$. \texttt{InfoLM} with $\mathcal{L}_\infty$ is closed to $0$ if $\forall i \in [1,|\Omega|] \quad p_i \approx q_i$. It means that all likely tokens (according to the \texttt{PMLM}) when considering $\pmb{x_i}$ are also likely when considering $\pmb{y_i^s}$. For string matching metrics, it corresponds to a \emph{perfect match} between $\pmb{x_i}$ and $\pmb{y_i^s}$. Second, let us consider $\pmb{x_i},\pmb{y_i^s}$ such that  $\mathcal{I}\left[p_{\mathbf{\Omega} | \mathbf{T}}(\cdot |\pmb{x_i}; \theta; \Tau),p_{\mathbf{\Omega} | \mathbf{T}}(\cdot |\pmb{y_i^s}; \theta; \Tau)\right] \approx 1$ (dissimilar texts) and a measure of information that relies on product of $p_i$ and $q_i$ (\textit{e.g} Fisher-Rao). In this case, $\forall i \in [1,|\Omega|] \quad p_i \times q_i \approx 0$ thus all likely tokens when considering $\pmb{x_i}$ are unlikely when considering $\pmb{y_i^s}$ (the converse it true as well). For string matching metrics this corresponds to \emph{no match} among the sub-strings of $\pmb{x_i}$ and $\pmb{y_i^s}$.

% Contrarily to methods that rely solely on continuous representation, computing $\mathcal{D}_r$ and $\mathcal{D}_c$ allow to interpret and gain insights on the distance between sentences.

\section{Experimental Frameworks}
In this section, we describe our experimental setting. We present the tasks and the baselines metrics use for each task.
\subsection{Text Summarization}
Text summarization aims at compressing long texts into fluent, short sentences that preserve the salient information. 
\\\textbf{Datasets.} To compare the different metrics previous work \cite{bhandari2020re} either relies on the TAC datasets \cite{dang2008overview,mcnamee2009overview} or on new summarization datasets extracted from CNN/DailyMail \cite{nallapati2016abstractive}. As pointed out by \citet{peyrard2019studying,bhandari2020re}, TAC datasets are old and contain flaws (\textit{e.g} systems used to generate summaries were of poor quality), we choose to work with the newly assemble dataset from CNN/Daily News proposed in \citet{bhandari2020re}. This dataset gathers 11,490 summaries and annotations are carried using the pyramid method \cite{nenkova2004evaluating}. 
\\\textbf{Metrics.} For text summarization, perhaps the most known metrics are \texttt{ROUGE} and its extensions \cite{ng2015better}, or \texttt{METEOR} and its variants \cite{denkowski2014meteor}. Recently, a new set of metrics (\textit{e.g} \texttt{BERTSCORE}, \texttt{MOVERSCORE}) have been applied to text summarization.
\subsection{Data2Text Generation} Prior works mainly rely on two task-oriented dialogue datasets (\textit{i.e.}, BAGEL \cite{mairesse2010phrase}, SFHOTEL \cite{wen2015semantically}). As sentence generated in these data-sets are unlikely to be representative of the progress of recent NLG systems we instead rely on a different dataset coming from the WebNLG2020 challenge \cite{gardent2017creating}. Given the following example of triple:  \textit{(John\_Blaha birthDate 1942\_08\_26) (John\_Blaha birthPlace San\_Antonio) (John\_E\_Blaha job Pilot)} the goal is to generate \textit{John Blaha, born in San Antonio on 1942-08-26, worked as a pilot}. 
\\\textbf{Annotations.} The WebNLG task is evaluated by human annotators along four different axes: (1)  Data Coverage: Are all the descriptions presented in the data included in the text? (2) Relevance: Does the text contains only predicates found in the data? (3) Correctness: Are predicates found in the data correctly mentioned and adequately introduced? (4) Text structure: Is the produced text well-structured, grammatically correct and written in acceptable English?, (5) Fluency: Does the text progress naturally? Is it easy to understand? Is it a coherent whole?
\\\textbf{Metrics.} For this task, organisers rely on untrained metrics (\textit{e.g.} \texttt{BLEU}, \texttt{METEOR}, \texttt{TER}, \texttt{BERTSCORE}) to compare the performance of the candidate systems. Thus, we will focus on system-level correlation. 	
%https://webnlg-challenge.loria.fr/files/WebNLG-2020-Presentation.pdf

\section{Numerical Results}

\begin{figure*}%\vspace{-0.5cm}
\centering
\resizebox{\textwidth}{!}{\begin{tabular}{cccc}
\subfloat[Abs - Text]{\includegraphics[trim=50 20 0 0,width = 1.05in]{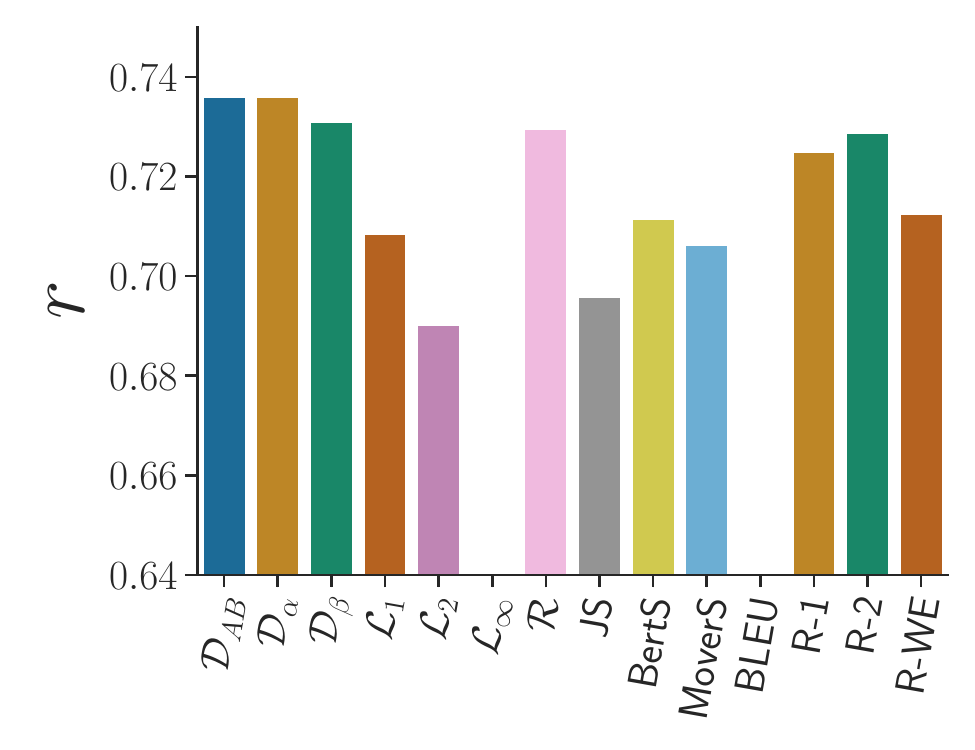}} &
\subfloat[Ext - Text]{\includegraphics[trim=0 20 0 0,width = 1.05in]{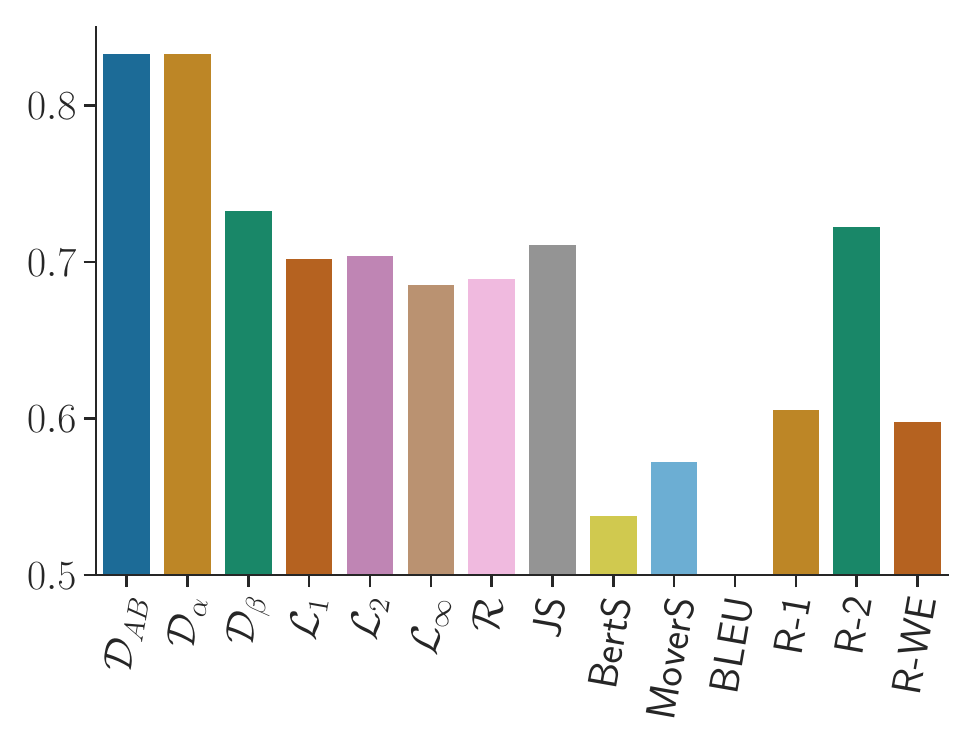}} &
\subfloat[Abs - Sys]{\includegraphics[trim=0 20 0 0,width = 1.05in]{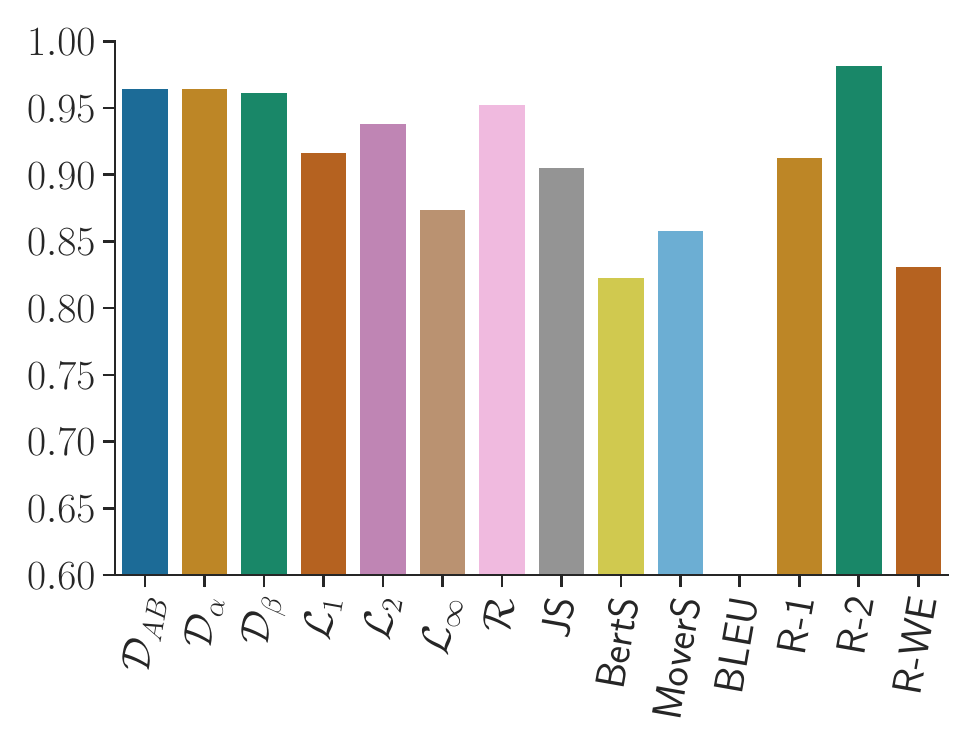}} &
\subfloat[Ext - Sys]{\includegraphics[trim=0 20 0 0,width = 1.05in]{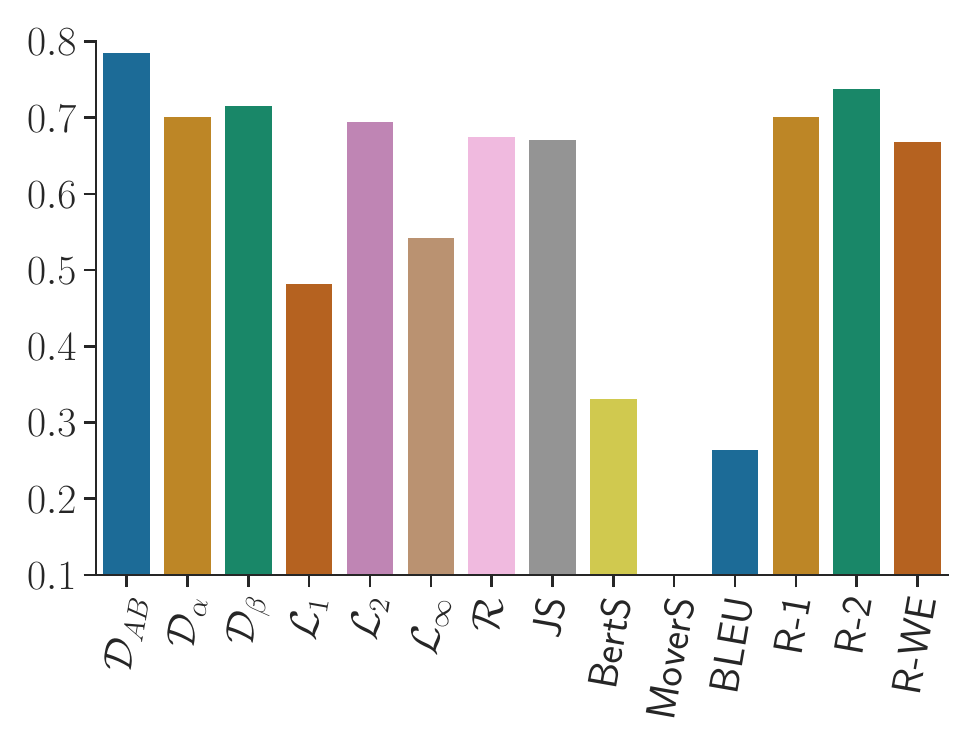}}\\
\subfloat[Abs - Text]{\includegraphics[trim=50 20 0 0,width = 1.05in]{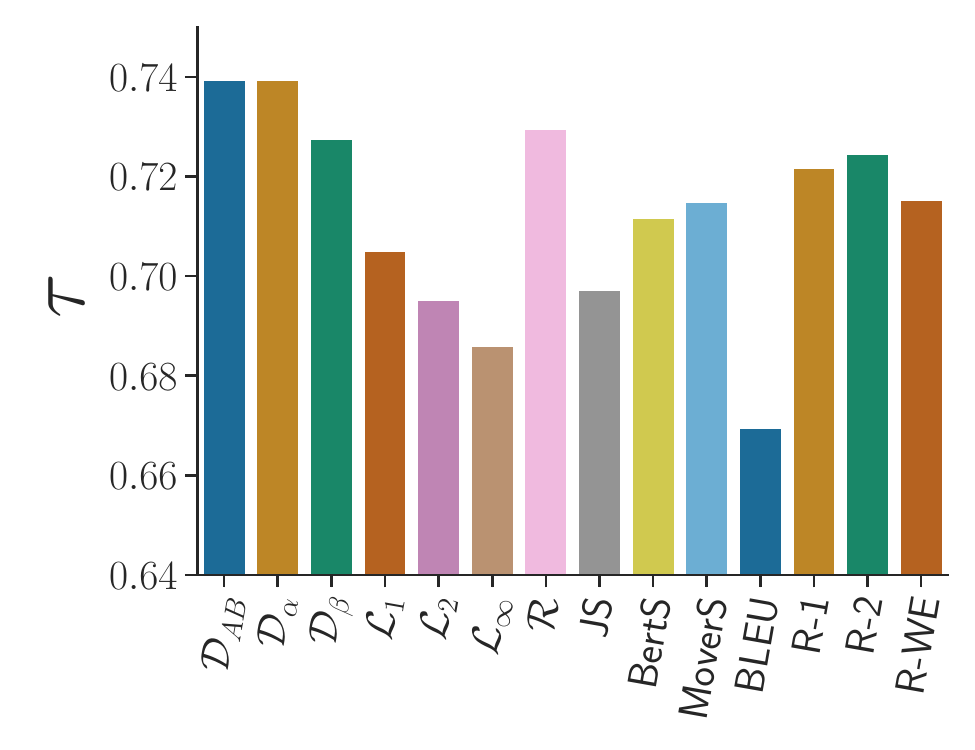}} &
\subfloat[Ext - Text]{\includegraphics[trim=0 20 0 0,width = 1.05in]{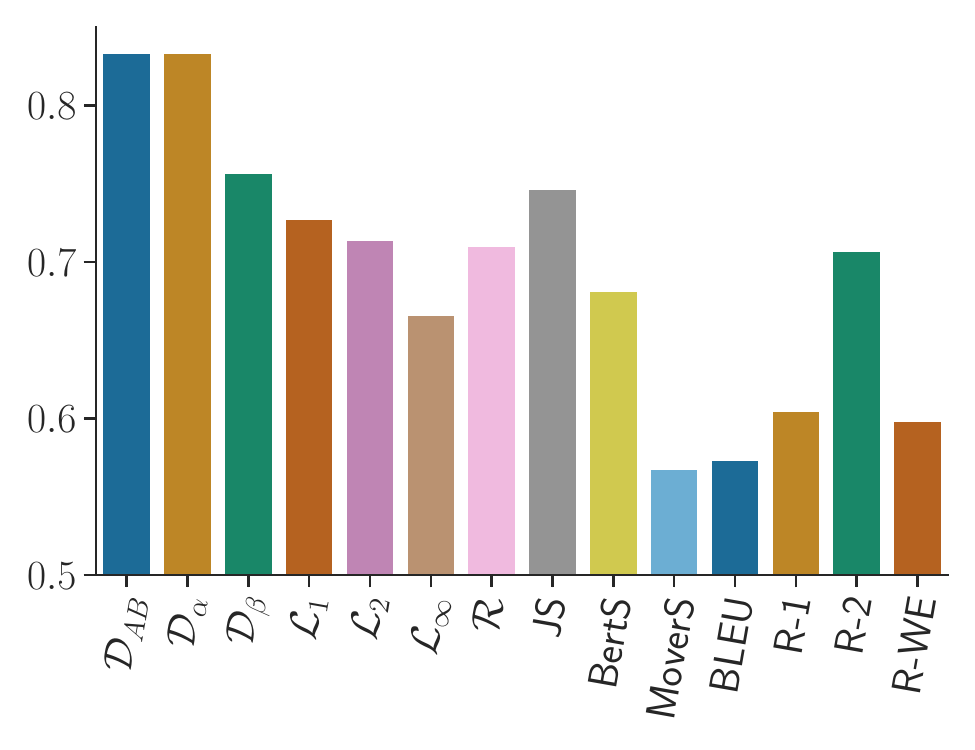}}&
\subfloat[Abs - Sys]{\includegraphics[trim=0 20 0 0,width = 1.05in]{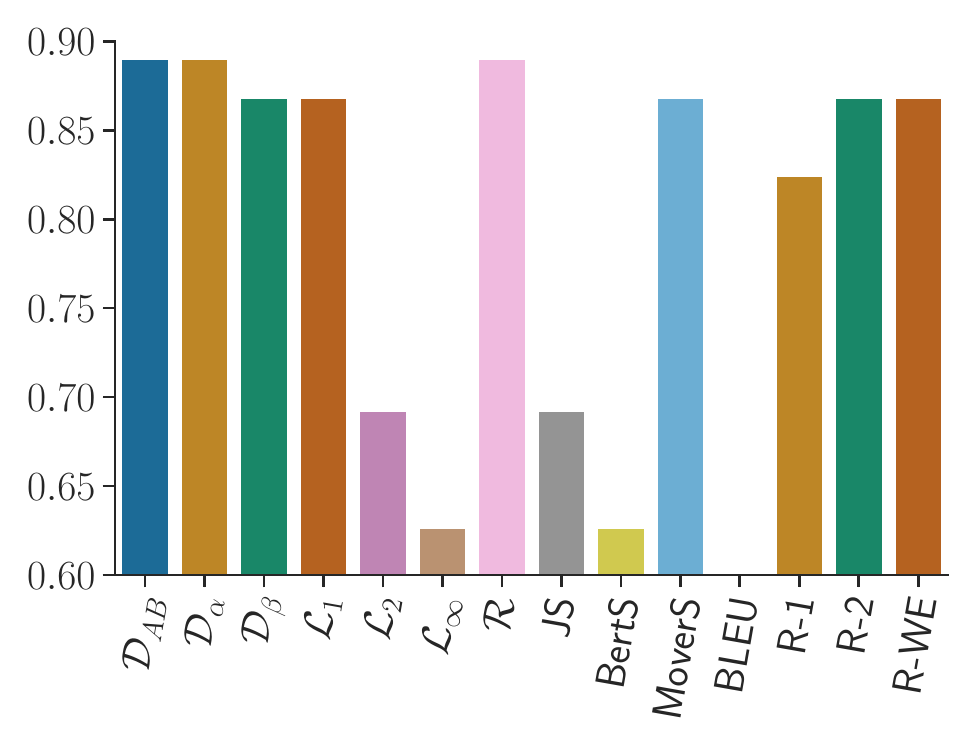}} &
\subfloat[Ext - Sys]{\includegraphics[trim=0 20 0 0,width = 1.05in]{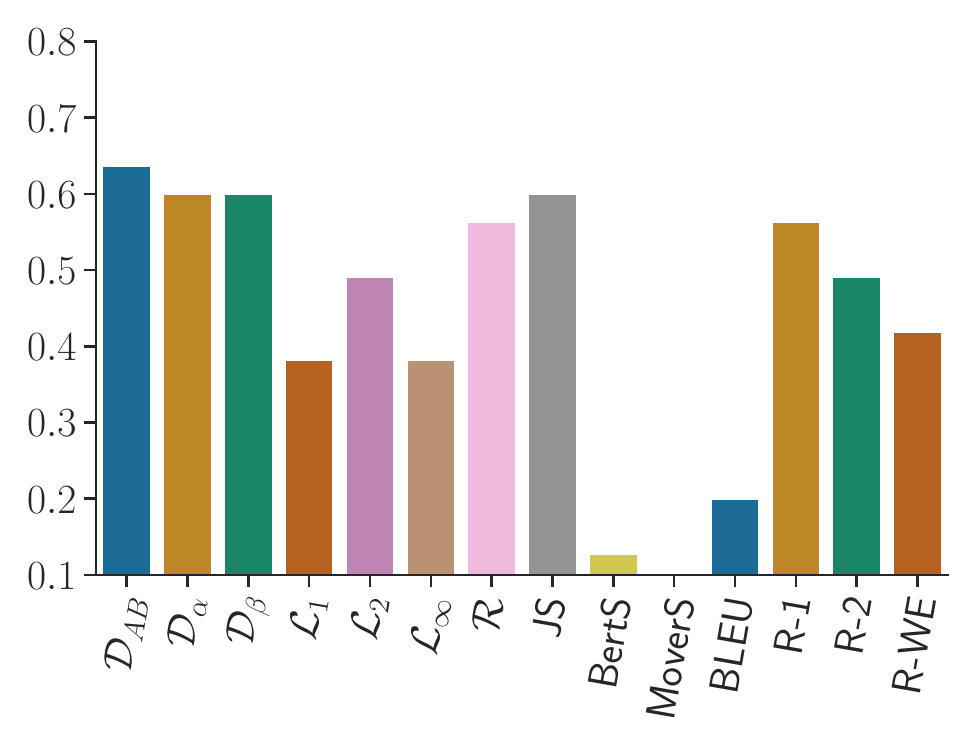}}\\
\end{tabular}}%\vspace{-.2cm}
\caption{Results of the correlation between metrics and human judgments on the CNN dataset. First raw reports correlation as measured by the Person ($r$) and second raw focus on Kendall ($\tau$) coefficient. In this experiment, parameter are optimized for each criterion.}\label{fig:cnn_results_all}
%The results on with Spearman ($\rho$) can be found in \ref{ssec:all_results_on_cnn_additional}).

% First, second and third row report correlations as measured by the Person ($r$), Spearman ($\rho$) and Kendall ($\tau$) coefficient respectively. %First, second and third, fourth columns report correlation coefficient at the text level (Text) where third and fourth columns report correlation at the system level (Sys). First and third column are dedicated to abstractive systems (Abs), second and fourth row report results for extractive systems. 
%\vspace{-.4cm}
\end{figure*}
In this section, we study the performance of \texttt{InfoLM} on both text summarization and data2text generation.

% In this section, we evaluate our metric \texttt{InfoLM} on a data2text task. In this challenge, automatic metrics are used to compare different systems. We first present a correlation analysis with the human judgement that we complete with statistical analysis to answer the following question: is \texttt{InfoLM} significantly better than the metrics used for comparing systems? , we then study the correlation matrix between the different metrics to better understand the difference of score between \texttt{InfoLM} and other standard metrics (\textit{e.g.},  BertScore, MoverScore, BLEU, Rouge). Finally, we study how changing the language model calibration affects the metric performance and how robust is $\mathcal{D}_{AB}$ to the choice of $\alpha$ and $\beta$.

\subsection{Results on Text Summarization}
\textbf{General Analysis}. \ref{fig:cnn_results_all} gathers the results of the correlation study between scores produced by different metrics and human judgement (\textit{i.e.} pyramid
score). We can reproduce results from \citet{bhandari2020re}. We observe a different behavior depending on the type of systems to be evaluated (\textit{e.g.},  abstractive or extractive) and the chosen correlation coefficient. We observe that \texttt{InfoLM} with $\mathcal{D}_{AB}$ or with $\mathcal{R}$ outperforms other BERT-based metrics such as \texttt{MOVERSCORE} or \texttt{BERTSCORE} (\textit{e.g.}, it is worth noting that both metrics perform poorly at the text or system-level when considering outputs from extractive systems). $\mathcal{D}_{AB}$ largely outperforms n-gram matching metrics (\textit{e.g.},  \texttt{ROUGE} metrics) on all datasets when measuring correlation with the Kendall $\tau$ and in almost all configurations (except when considering abstractive outputs at the system level) when using the Pearson $r$. It is worth noting the overall good performance of the parameter-free Fisher-Rao distance.
\\\textbf{Choice of information geometric measure for \texttt{InfoLM}.} In \ref{fig:cnn_results_all}, we can observe two different types of groups depending on the global behaviour. First we notice that using $\mathcal{L}_p$, $p \in \{1,2,\dots, +\infty\}$ leads to poor performances in many configurations. Good performance of $\mathcal{L}_\infty$ in some configurations is surprising as $\mathcal{L}_\infty$ is extremely selective  (\textit{i.e.} $\mathcal{L}_\infty$ computes $\max_i|p_i - q_i|$). As output produced by the \texttt{PMLM} is sparse, $\max_i|p_i - q_i|$ correspond to one likely word in one sentence and not likely at all in the other. The second group gathers $JS$,  $\mathcal{R}$, $\mathcal{D}_\alpha$, $\mathcal{D}_\beta$ and $\mathcal{D}_{AB}$ and achieves the best performance overall. $\mathcal{D}_\alpha$ and $\mathcal{D}_{AB}$ achieve similar performance suggesting that the flexibility (\textit{e.g.},  robustness to outliers) introduced by the $\beta$ parameter in $\mathcal{D}_{AB}$ is not useful in our task. This observation is strengthened by the lower performance of $\mathcal{D}_{\beta}$. The difference of results between the two measures is due to the flexibility introduced by $\alpha$ (\textit{i.e.}, $\alpha$ controls the relative importance of the ration $\frac{p_i}{q_i}$) which can be interpreted in our case as the ability to control the importance attributed to less likely words \cite{hamid}. 
\\\textbf{Takeaways.} The best performing metric is obtained with $\mathcal{D}_{AB}$. The Fisher-Rao distance, denoted by $\mathcal{R}$, achieves good performance in many scenarios and has the advantage to be parameter-free. 

% We can observe groupes of metric that correlates. a/b divergences exhibit similarity in the formula.
% LP are of the same familly but.... Linfinity is allone + bleue
% Rouge metric as well 
% Are of different nature. Choosing the geometry affect the nature which demonstrate both the generality and the flexibility of our approach.

% State that we observe onb sys level because it is how it is used to compare system in the web nlg task.
\subsection{Results on Data2Text}
\textbf{Global Analysis:} \ref{tab:web_nlg_sys} gathers results of the correlation analysis of the metrics with human judgements following the five different axes. We observe that the five considered criteria of annotations are not independent: text structure and fluency achieve a strong correlation coefficient ($>98$). Additionally, all metrics achieve similar results when the correlation is computed on these two criteria. We observe that the best performing group of metric is based on \texttt{InfoLM} followed by metrics based on continuous representation from BERT (\textit{i.e.}, \texttt{MOVERSCORE} and \texttt{BERTSCORE}) followed by N-gram matching metrics.
Regarding correctness, data coverage and relevance, we observe that both $\mathcal{D}_{AB}$ and $\mathcal{D}_{\alpha}$ achieve the best results on almost all correlation coefficients. On data coverage, \texttt{InfoLM} achieves improvement up to $17$ points in correlation compared to both BERT based or N-gram matching metrics. Regarding fluency and text structure, Fisher-Rao distance works better and slightly outperforms the second-best performing metric, namely \texttt{BERTSCORE}.
\\\textbf{Takeaways.} Similar to summarisation, we observe very low correlation for ${L}_p$, $p \in \{1,2,\dots,+\infty \}$. We also observe that $\beta$-divergences achieve lower results than both $\alpha$ and $AB$ divergences suggesting that, as noticed for summarisation, robustness to unlikely words (\textit{i.e.}, outliers) is less relevant for our task. 

\begin{table*}[]
    \centering

 \resizebox{\textwidth}{!}{\begin{tabular}{l|rrr|rrr|rrr|rrr|rrr|}\hline  
   &  \multicolumn{3}{c}{Correctness} & \multicolumn{3}{c}{Data Coverage} & \multicolumn{3}{c}{Fluency} &  \multicolumn{3}{c}{Relevance} &   \multicolumn{3}{c}{Text Structure} \\
  Metric &  $r$ &  $\rho$ & $\tau$  &  $r$ &  $\rho$ &  $\tau$ & $r$ &  $\rho$ &  $\tau$  &  $r$ &  $\rho$ &  $\tau$  & $r$ &  $\rho$ &  $\tau$  \\
\midrule               Correct &   100.0    &    100.0    &    100.0   &    97.6 &     85.2 &    73.3 &    80.0 &     81.1 &    61.6 &    99.1 &     89.7 &    75.0 &    80.1 &     80.8 &    60.0 \\                 
DataC &      85.2 &       97.6 &      73.3 &  100.0   &    100.0  &   100.0  &    71.8 &     51.7 &    38.3 &    96.0 &     93.8 &    81.6 &    71.6 &     51.4 &    36.6 \\               
Fluency &      81.1 &       80.0 &      61.6 &    71.8 &     51.7 &    38.3 &  100.0   &  100.0    &  100.0   &    77.0 &     61.4 &    46.6 &    99.5 &     99.7 &    98.3 \\                
Relev &      89.7 &       99.1 &      75.0 &    96.0 &     93.8 &    81.6 &    77.0 &     61.4 &    46.6 &  100.0   &  100.0    &  100.0   &    77.2 &     61.1 &    45.0 \\                 
TextS &      80.8 &       80.1 &      60.0 &    71.6 &     51.4 &    36.6 &    99.5 &     99.7 &    98.3 &    77.2 &     61.1 &    45.0 &  100.0   &  100.0    &   100.0  \\   \hline
$ \mathcal{D}_{AB}$ &      {88.8} &       \underline{\textbf{89.3}} &      \underline{\textbf{76.6}} &    \underline{\textbf{81.8}} &     \underline{\textbf{82.6}} &    \underline{\textbf{70.0}} &    86.6 &     92.0 &    76.6 &    \underline{\textbf{89.8}} &     \underline{\textbf{87.9}} &    \underline{{73.3}} &    86.6 &     91.4 &    75.0 \\ 
$ \mathcal{D}_\alpha$ &      {88.8} &       \underline{\textbf{89.3}} &      \underline{\textbf{76.6}} &    \underline{\textbf{81.8}} &     \underline{\textbf{82.6}} &    \underline{\textbf{70.0}} &    86.6 &     92.0 &    76.6 &   \underline{ \textbf{89.8}} &     \underline{\textbf{87.9}} &    \underline{{73.3}} &    86.6 &     91.4 &    75.0 \\   
$ \mathcal{D}_\beta$ &      81.4 &       50.0 &      71.6 &    48.4 &     79.7 &    65.0 &    44.8 &     84.7 &    76.6 &    49.3 &     72.3 &    60.0 &    48.0 &     83.8 &    75.0 \\      
$ \mathcal{L}_1$ &      75.2 &       33.8 &      61.6 &    32.4 &     53.8 &    40.0 &    22.7 &     83.5 &    73.3 &    32.2 &     57.9 &    45.0 &    25.6 &     83.2 &    71.6 \\      
$ \mathcal{R}$ &      \underline{\textbf{89.7}} &       86.0 &      75.0 &    78.7 &     70.5 &    51.6 &    \underline{\textbf{93.3}} &     \underline{\textbf{95.7}} &    \underline{\textbf{85.3}} &   87.6 &     84.4 &    70.0 &    \underline{\textbf{92.4}} &     \underline{93.8} &    \underline{81.6} \\                   
JS &      79.4 &       81.1 &      70.0 &    69.3 &     75.5 &    60.0 &    89.4 &     91.4 &    75.0 &    81.7 &     70.5 &    60.0 &    91.9 &     91.1 &    73.3 \\ \hline                
BertS &      \underline{85.5} &       83.4 &      \underline{73.3} &    
74.7 &     \underline{68.2} &    53.3 &    
\underline{92.3} &     \underline{95.5} &    \underline{85.0} &    
\underline{83.3} &     \underline{79.4} &    \underline{65.0} &    
\underline{91.9} &     \underline{\textbf{95.0}} &    \underline{\textbf{83.3}} \\       
MoverS &      84.1 &       \underline{84.1} &      \underline{73.3} &    
\underline{78.7} &     66.2 &    \underline{53.3} &    
91.2 &     92.1 &    78.3 &    
82.1 &     77.4 &    65.0 &    
90.1 &     91.4 &    76.3 \\\hline  
BLEU &      77.6 &       66.3 &      60.0 &    
55.7 &     50.2 &    36.6 &    
\underline{89.4} &     90.5 &    78.3 &    
63.0 &     65.2 &    51.6 &    
{88.5} &     89.1 &    76.6 \\
R-1 &      80.6 &       65.0 &      65.0 &    61.1 &     \underline{59.6} &    \underline{48.3} &    
76.5 &     76.3 &    60.3 &    
64.3 &     \underline{69.2} &    56.7 &    
75.9 &     77.5 &    58.3 \\
METEOR &      \underline{86.5} &       \underline{66.3} &      \underline{70.0} &    
\underline{77.3} &     50.2 &    46.6 &    
86.7 &     90.5 &    78.3 &    
\underline{82.1} &     65.2 &    58.6 &    
86.2 &     89.1 &    76.6 \\
TER &      79.6 &       78.3 &      58.0 &    69.7 &     58.2 &    38.0 &    89.1 &     \underline{93.5} &    \underline{80.0} &    75.0 &     70.2 &    \underline{\textbf{77.6}} &    \underline{89.5} &     \underline{91.1} &    \underline{78.6} \\
\bottomrule\end{tabular}}
\caption{Correlation at the system level with human judgement along five different axis: correctness, data coverage, fluency, relevance and text structure for the WebNLG task. Best results by group are underlined, overall best results are bolted.}
    \label{tab:web_nlg_sys}
\end{table*}
\subsection{Further Analysis}
\subsubsection{Correlation Between Metrics}
\begin{figure}[!htb]
       \centering
    \includegraphics[width=0.35\textwidth]{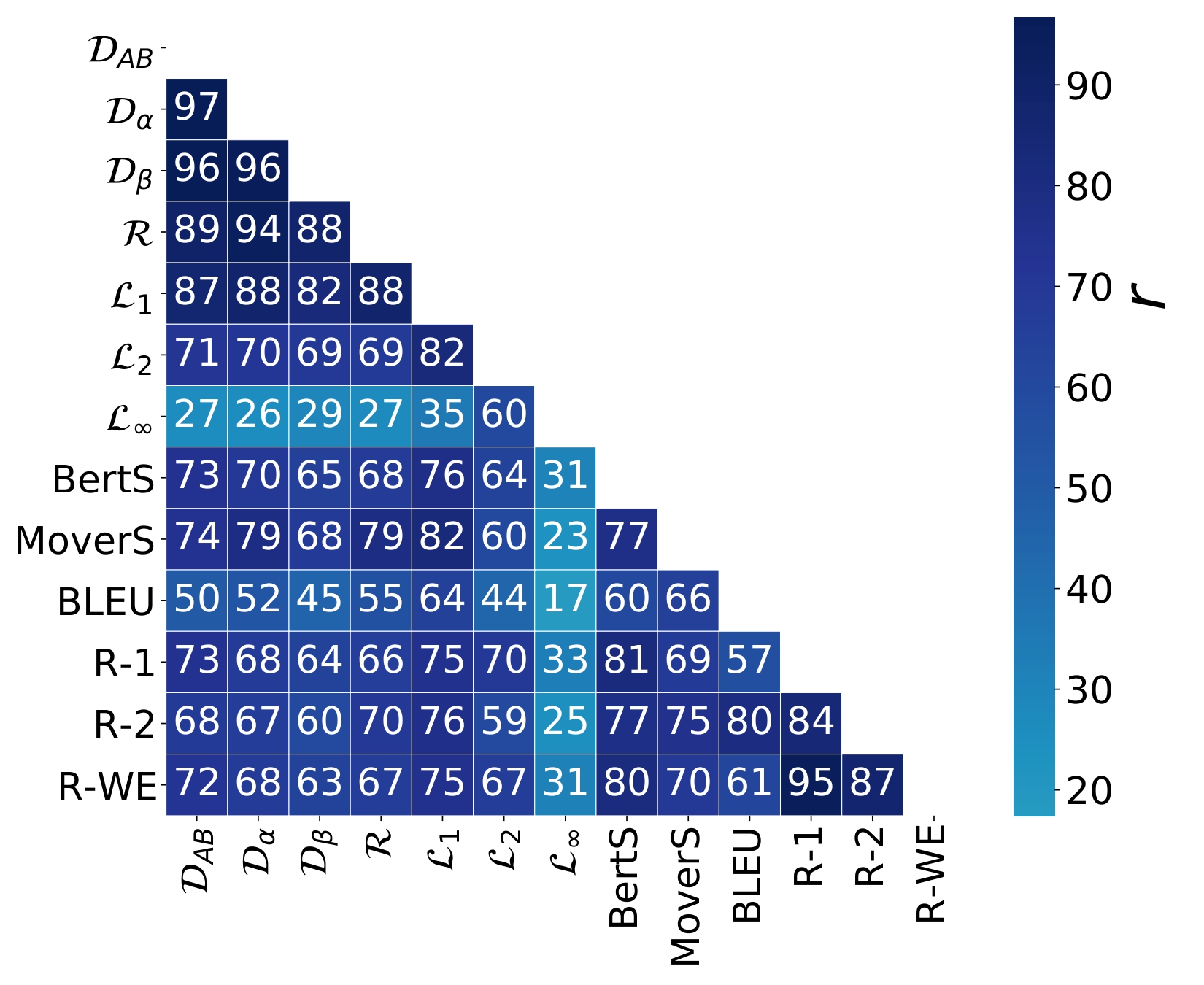}
    \caption{Pearson correlation at the system level between metrics when considering abstractive system outputs.}
    \label{fig:correlation_metrics}
\end{figure}

In this experiment, we complete our global analysis by comparing the scores obtained by the different metrics with each other. We want to gain an understanding of how different our metric is from other metrics and how the choice of information geometric measures affects the predictions. \ref{fig:correlation_metrics} gathers the results of the experiment. We observe a high correlation ($r > 88$) between $\mathcal{D}_\alpha$, $\mathcal{D}_\beta$, $\mathcal{D}_{AB}$ and $\mathcal{R}$\footnote{Note that these metrics consider the product of $p_i$ and $q_i$.}. Interestingly, we observe a lower correlation ($r \approx 70$) with  \texttt{BERTSCORE} and N-gram matching metrics, e.g.,  \texttt{ROOGE}) whereas \texttt{BERTSCORE} achieves a stronger correlation with \texttt{ROUGE} ($r \approx 80$).
% insight on how similar are the geométries :)
\\\textbf{Takeaways.} Through the correlation analysis in \ref{fig:correlation_metrics}, we observe the impact of different geometry on \texttt{InfoLM} predictions. The correlation analysis shows that the prediction of \texttt{InfoLM} when using $\mathcal{D}_\alpha$, $\mathcal{D}_\beta$, $\mathcal{D}_{AB}$ and $\mathcal{R}$ are highly correlated and as illustrated by previous experience achieve high correlation scores which we believe validate our approach. It is worth noting that $\mathcal{R}$ requires no tuning as it is parameter free.

\subsubsection{Score Distributions}
\begin{figure}[!htb]
       \centering
    \includegraphics[width=0.47\textwidth]{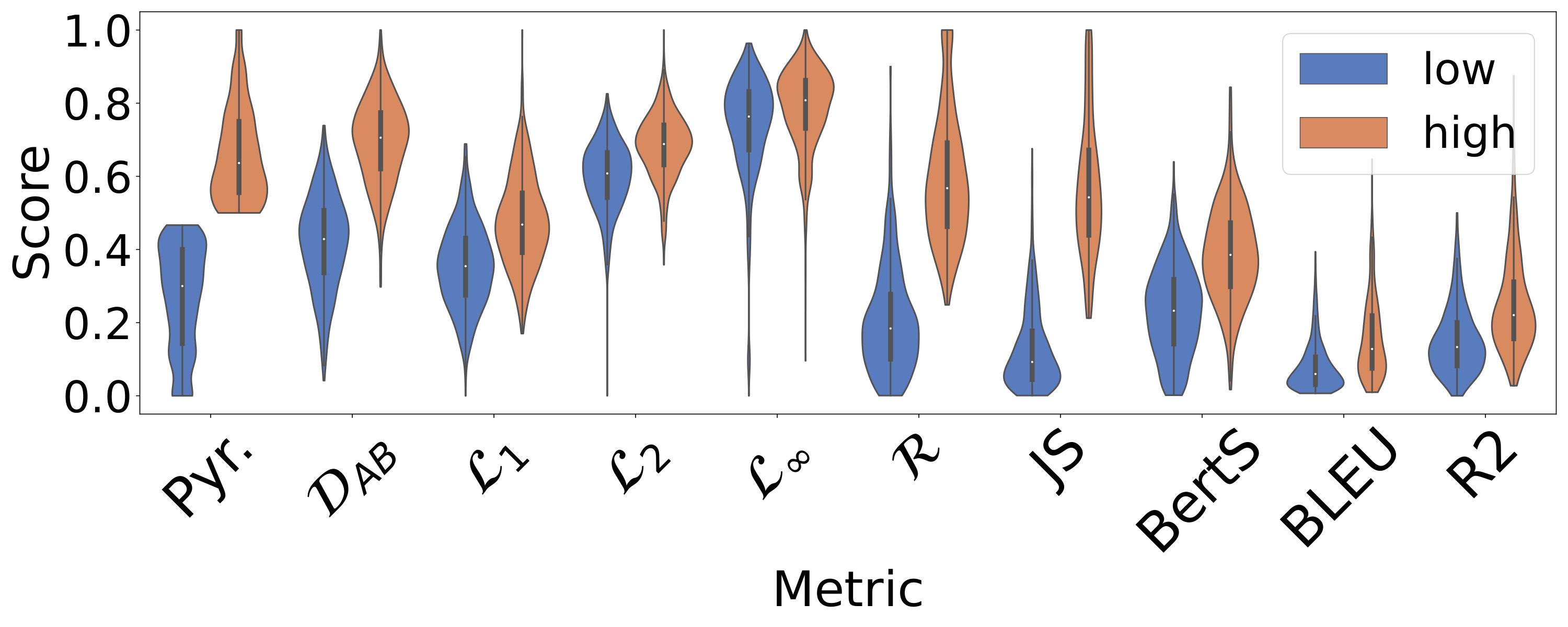}
    \caption{Score distribution of text score when considering abstractive system outputs. $Pyr.$ stands for pyramide score.}
    \label{fig:score_distrib_small}
\end{figure}

In \ref{fig:score_distrib_small}, we study the text score distribution of different metrics on abstractive summary. The ideal metric would mimic the human score distribution (\textit{i.e.} $Pyr.$) and be able to distinguish between good and bad quality summaries. The results show that \texttt{ROUGE} and \texttt{BLEU} struggle to distinguish between between good quality ($Pyr. \geq 0.5$) low quality summaries ($Pyr. \leq 0.5$) which has been reported in \citet{peyrard2019studying}. We observe that $\mathcal{D}_{AB}$, $\mathcal{R}$ and $JS$ metrics are able to make the distinction. Interestingly, as $p$ increases and the $\mathcal{L}_p$ distances become more selective (\textit{i.e.} focus on one word solely), the $\mathcal{L}_p$ distances struggle to distinguish low from high scoring summaries. 
\\\noindent\\\textbf{Takeaways.} \texttt{InfoLM} when combined with $\mathcal{D}_{AB}$, $\mathcal{R}$ and $JS$ is able to distinguish high-scoring from low scoring summaries.

\subsubsection{Temperature Calibration}
\begin{figure}[!htb]
       \centering
    %\hspace{-.5cm}
    \includegraphics[width=0.3\textwidth]{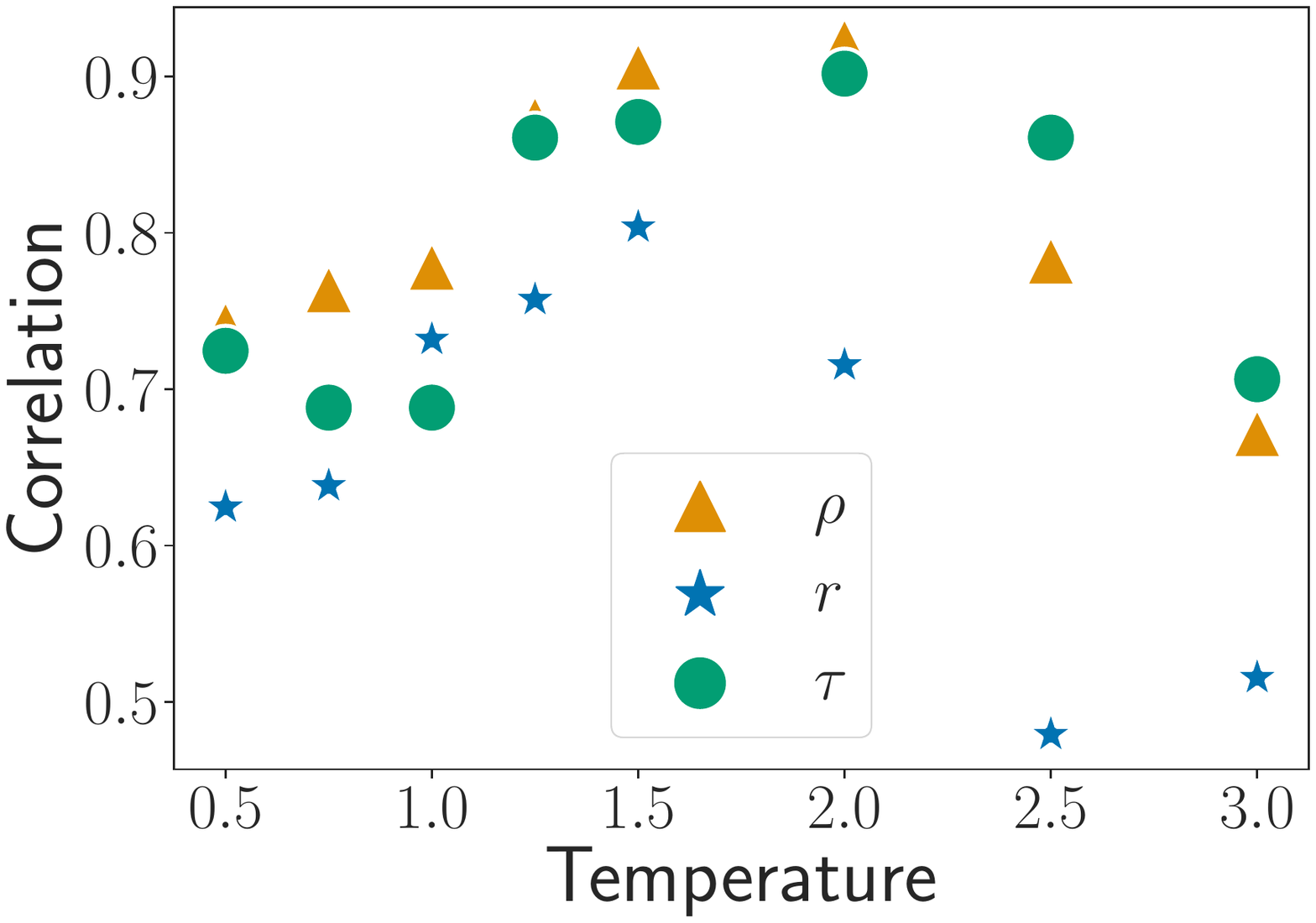}\caption{Impact of Calibration %on system-level correlation 
    for summarization.}\label{fig:calibration_rao_main} %\vspace{-0.5cm}}
\end{figure}

To study the impact of calibration, we choose to work on system-level correlation and report in  \ref{fig:calibration_rao_main} the achieved the correlation measured by the different coefficients. We limit our study to the Fisher-Rao distance as it is a parameter-free metric and is among the best-performing metrics of \texttt{InfoLM}. Due to space constraints, we report the result on extractive systems only. %(extractive systems can be found in  \ref{fig:calibration_rao_all_total}). 
\\\noindent\textbf{Takeaways.} Fisher-Rao only considers product $p_i\times q_i$ thus as $\mathcal{T}$ increases and the predicted probability of the \texttt{PMLM} becomes more uniform more words are considered and the aggregated distributions become richer in terms of considered words. It is worth noting that when changing the temperature we observe a smooth change in correlation and observe an optimal temperature $T$ which is reached for $T \in [1,2]$. It suggests that \texttt{InfoLM} benefits from a \texttt{PMLM} that is not too selective (case $T \ll 1$).  For a specific application, the temperature of \texttt{InfoLM} can be tuned to improve correlation and \texttt{InfoLM} will likely benefit from well-calibrated \texttt{PMLM}.

\section{Summary and Concluding Remarks}
In this work, we presented \texttt{InfoLM} that does not require training and it is among the first metrics computing the similarity between two discrete probability distributions over the vocabulary (which is similar to string-based metrics) but also leverages the recent advance in language modeling thanks to a \texttt{PMLM}. Our experiments on both summarization and data2text generation demonstrate the validity of our approach. Among available contrast measures, the Fisher-Rao distance is parameter-free and thus, it is easy to use in practice while the $AB$-Divergence achieves better results but requires to select $\alpha$ and $\beta$. 
Future work includes extending our metrics to new tasks such as SLU \cite{chapuis2020hierarchical,chapuis2021code,dinkar2020importance,colombo2021novel}, controlled sentence generation \cite{colombo2019affect,colombo2021beam} and multi-modal learning \cite{colombo2021improving,garcia-etal-2019-token}.

\section*{Acknowledgments}
This work was also granted access to the HPC resources of IDRIS under the allocation 2021-AP010611665 as well as under the project 2021-101838 made by GENCI. 

%\newpage
\bibliography{bib.bib}

\appendix

%\include{appendix}
%\bibliography{bib.bib}

\end{document}